\documentclass{article}

\usepackage{arxiv}

\usepackage[utf8]{inputenc} % allow utf-8 input
\usepackage[T1]{fontenc}    % use 8-bit T1 fonts
\usepackage{hyperref}       % hyperlinks
\usepackage{url}            % simple URL typesetting
\usepackage{booktabs}       % professional-quality tables
\usepackage{amsfonts}       % blackboard math symbols
\usepackage{nicefrac}       % compact symbols for 1/2, etc.
\usepackage{microtype}      % microtypography
\usepackage{graphicx}
\usepackage{natbib}
\usepackage{doi}
\usepackage{amsmath} % for mathematical symbols and equations
\usepackage{graphicx} % for including graphics
\usepackage{pifont}
\usepackage{multirow}
\usepackage{xcolor}

\title{Separating Invisible Sounds Toward \\ Universal Audiovisual Scene-Aware Sound Separation}

\date{} 					% Or removing it
\author{ 
    Yiyang Su\thanks{Equal contribution.} \\
    Michigan State University \\
    \texttt{suyiyan1@msu.edu} \\
    \And
    Ali Vosoughi$^*$ \\
    University of Rochester \\
    \texttt{mvosough@ur.rochester.edu} \\
    \And
    Shijian Deng$^*$ \\
    University of Texas at Dallas \\
    \texttt{shijian.deng@utdallas.edu} \\
    \And
    Yapeng Tian \\
    University of Texas at Dallas \\
    \texttt{yapeng.tian@utdallas.edu} \\
    \And
    Chenliang Xu \\
    University of Rochester \\
    \texttt{chenliang.xu@rochester.edu} \\
}

\renewcommand{\headeright}{ICCV 2023 - AV4D}
\renewcommand{\undertitle}{accepted at ICCV 2023 - AV4D}
\renewcommand{\shorttitle}{AVSA-Sep}

\usepackage{xspace} % for handling spacing

\newcommand{\onedot}{.}
\newcommand{\etal}{\emph{et al}\onedot}
\newcommand{\vs}{\emph{vs}\onedot}

\hypersetup{
pdftitle={Separating Invisible Sounds Toward \\ Universal Audiovisual Scene-Aware Sound Separation},
pdfsubject={Computer Vision, Audio, Audivisual},
}

\begin{document}
\maketitle

\begin{abstract}
The audio-visual sound separation field assumes visible sources in videos, but this excludes invisible sounds beyond the camera's view. Current methods struggle with such sounds lacking visible cues. This paper introduces a novel \textit{\textbf{A}udio-\textbf{V}isual \textbf{S}cene-\textbf{A}ware \textbf{Sep}aration} (AVSA-Sep) framework. It includes a semantic parser for visible and invisible sounds and a separator for scene-informed separation. AVSA-Sep successfully separates both sound types, with joint training and cross-modal alignment enhancing effectiveness.
\end{abstract}

% keywords can be removed
% \keywords{First keyword \and Second keyword \and More}

\section{Introduction}

Recent progress in audiovisual sound separation (AVSS) enables enhanced separation using aligned audiovisual cues~\cite{tian2021cyclic, liu2022separate, tzinis2022audioscopev2,zhu2020visually, zhao_sound_2018, Afouras20b, gao2019co, majumder2021move2hear, tzinis2020into}. Yet challenges arise when visual cues are absent, as in off-screen narrations or close-ups. We categorize sources as \textit{visible sound} (in the visual scope) and \textit{invisible sounds} (outside the visual scope).

Vital for sound separation is addressing invisible sounds. They are common in videos because most cameras have a limited field of view. Existing AVSS methods~\cite{zhao_sound_2018, xu2019recursive, tian2021cyclic} can predict the difference between the sound mixture and the visible sounds as the invisible sound, yet they can not deal with multiple invisible sources.

We propose a novel audiovisual sound separation framework, \textbf{A}udio\textbf{V}isual \textbf{S}cene-\textbf{A}ware \textbf{Sep}aration (AVSA-Sep), which leverages video scene semantics as a substitute for visual cues. We contend that AVSS techniques leverage visual semantics for guiding sound separation. Hence, we incorporate audio semantics when visual semantics are absent.

\begin{figure}[tbp]
    \centering
    \includegraphics[width=0.7\linewidth]{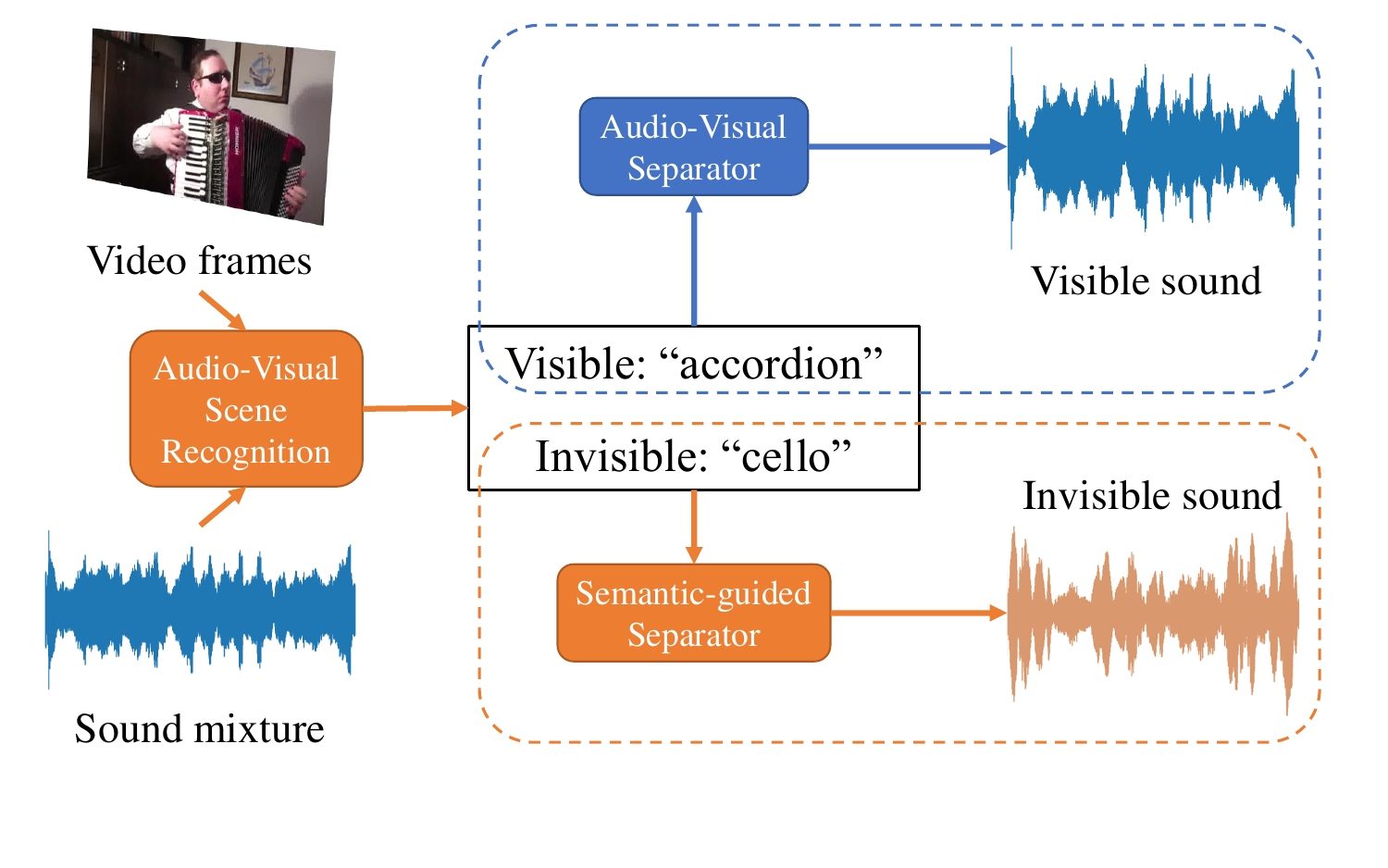}
    \vspace{-2em}
    \caption{
    The AVSA-Sep framework. It predicts visible and invisible scenes from frames and sound mixture, then separates sounds. The audiovisual separator estimates visible sounds, while the semantic-guided one estimates invisible sounds. This separates both visible (e.g., blue waveform's accordion) and invisible (e.g., orange waveform's cello) sound sources.
    }
    \label{fig:overview}
\end{figure}

As illustrated in Fig.~\ref{fig:overview}, in our approach, we begin with audiovisual scene recognition to grasp scene-level semantics from the video. Next, an audiovisual separator predicts visible sounds, while a semantic-guided separator predicts invisible sounds. This scene-aware setup can handle both visible sounds and more than one invisible sound source.
% Ablation studies expose a performance bottleneck and confirm our effective joint training.

Our contributions are threefold: 1) addressing invisible sounds, especially multiple invisible sounds, in audiovisual context; 2) introducing AVSA-Sep that leverages video scene semantics for visible and invisible sound separation; 3) incorporating semantic parsing into our framework, which helps to separate invisible sounds. 

%------------------------------------------------------------------------
\section{Related Work}

\paragraph{Blind source separation (BSS).}
In audio signal processing, BSS methods untangle mixtures into source signals without additional cues like visuals. Classical approaches, such as Independent Component Analysis~\cite{hyvarinen2000independent}, Principal Component Analysis~\cite{huang2012singing}, and Non-negative Matrix Factorization~\cite{lee2000algorithms, virtanen2007monaural, ozerov2009multichannel}, exploit statistical properties for source independence. Spatial audio-based methods~\cite{fitzgerald2016projet, rickard2007duet, fitzgerald2016projection} utilize location information. Deep-learning-based methods~\cite{hershey2016deep, wang2018alternative, yu2017permutation, kolbaek2017multitalker} leverage deep networks to capture features. As shown in Tab.~\ref{tab:scenarios}, while some BSS methods tackle multiple invisible sounds, audiovisual synergy is absent.

\paragraph{Audiovisual sound separation.}
AVSS models have gained momentum since the inception of Sound of Pixels (SoP)~\cite{zhao_sound_2018}. Subsequent works~\cite{xu2019recursive, gao2018learning, tzinis2022audioscopev2, tzinis2020into} have further harnessed audiovisual correspondences. Various aspects, such as motion~\cite{zhao_sound_2019}, gestures~\cite{gan_music_2020, rahman2021tribert}, natural language~\cite{rahimi2022reading, tan2023language}, embodied AI~\cite{majumder2022active, majumder2021move2hear}, object localization~\cite{gao2019co}, visual grounding~\cite{tian2020co, tian2021cyclic}, speech separation~\cite{lee2021looking, gao2021visualvoice, montesinos2022vovit}, and scene graphs~\cite{chatterjee2022learning, chatterjee2021visual}, have been incorporated. \cite{owens_audio-visual_2018} considers off-screen sounds but is limited to one invisible sound. Similar to iQuery~\cite{chen2023iquery}, we leverage semantic labels. While other AVSS methods handle up to one invisible sound, we address the challenge of separating multiple invisible sounds.

\begin{table}
\centering
\resizebox{\linewidth}{!}{
\begin{tabular}{lcccc}
\toprule
& \textbf{BSS} & \textbf{AVSS} & \textbf{AVSA-Sep} \\
% \textbf{} & (e.g., RPCA) & (e.g., SoP) & (Ours) \\
\textbf{} & (e.g.,~\cite{huang2012singing}) & (e.g.,~\cite{zhao_sound_2018}) & (Ours) \\
\midrule
Uses visual cues & \textcolor{red}{\ding{55}} & \textcolor{green}{\ding{51}} & \textcolor{green}{\ding{51}} \\
2+ invisible sounds & Some & \textcolor{red}{\ding{55}} &  \textcolor{green}{\ding{51}} \\
\bottomrule
\end{tabular}
}
\caption{BSS methods, such as RPCA~\cite{huang2012singing}, lack visual cues and sometimes handle up to 2 sounds. AV-Sep methods like SoP~\cite{zhao_sound_2018} address only one invisible sound, while our approach benefits from visual cues and can separate multiple invisible sounds.}
\label{tab:scenarios}
\end{table}

\paragraph{Audiovisual scene understanding.}
Leveraging extensive audiovisual datasets \cite{gemmeke2017audio, vedaldi2020vggsound, lee2021acav100m} has driven a surge in audiovisual learning \cite{lin2023vision, gao2023collecting, gong2022contrastive, zhou2022audio, mercea2022temporal, lee2022audio, cheng2022joint, lee2022weakly, zellers2022merlot, vasudevan2022sound, afouras2022self, mercea2022audio, mo2022multi, mittal2022learning}. While these methods assume audiovisual correspondences, we incorporate semantics and representations of invisible sounds. Our approach further employs semantic information to guide the separation of invisible sounds.

%-------------------------------------------------------------------------
\section{Method}

% \subsection{Overview} \label{sec:sa-sep}

In the universal AVSS task, the goal is to recover individual sound sources ($S_1, S_2, \dots, S_m$) from visual frames ($I_1, I_2, \dots, I_n$) of $n$ visible sounds and the sound mixture ($S_{\text{mix}}$) of $m$ individual sounds. Traditional models assume a one-to-one correspondence ($m=n$), while our approach accommodates any $m$ and $n$. For clarity, we assume each frame $I_j$ corresponds to a sound $S_i$, as sounding object localization has been addressed by Tian~\etal~\cite{tian2020co}.

We introduce the AVSA-Sep framework to address this challenge. It leverages video scenes (frames or semantics) as intermediaries between input audio and separated sounds, as depicted in Fig.~\ref{fig:overview}. The framework consists of two steps: 1) a semantic parser predicting scenes from visual frames and audio mixture and 2) sound separators separating sound components from the mixture conditioned on either visual frames or semantic labels.
For clarity, we intentionally maintain simple architectures, anticipating that adapting newer AVSS baselines to our framework will yield improved performance.

\subsection{Semantic-Guided Sound Separation}

\begin{figure}[tbp]
    \centering
    \includegraphics[width=0.7\textwidth]{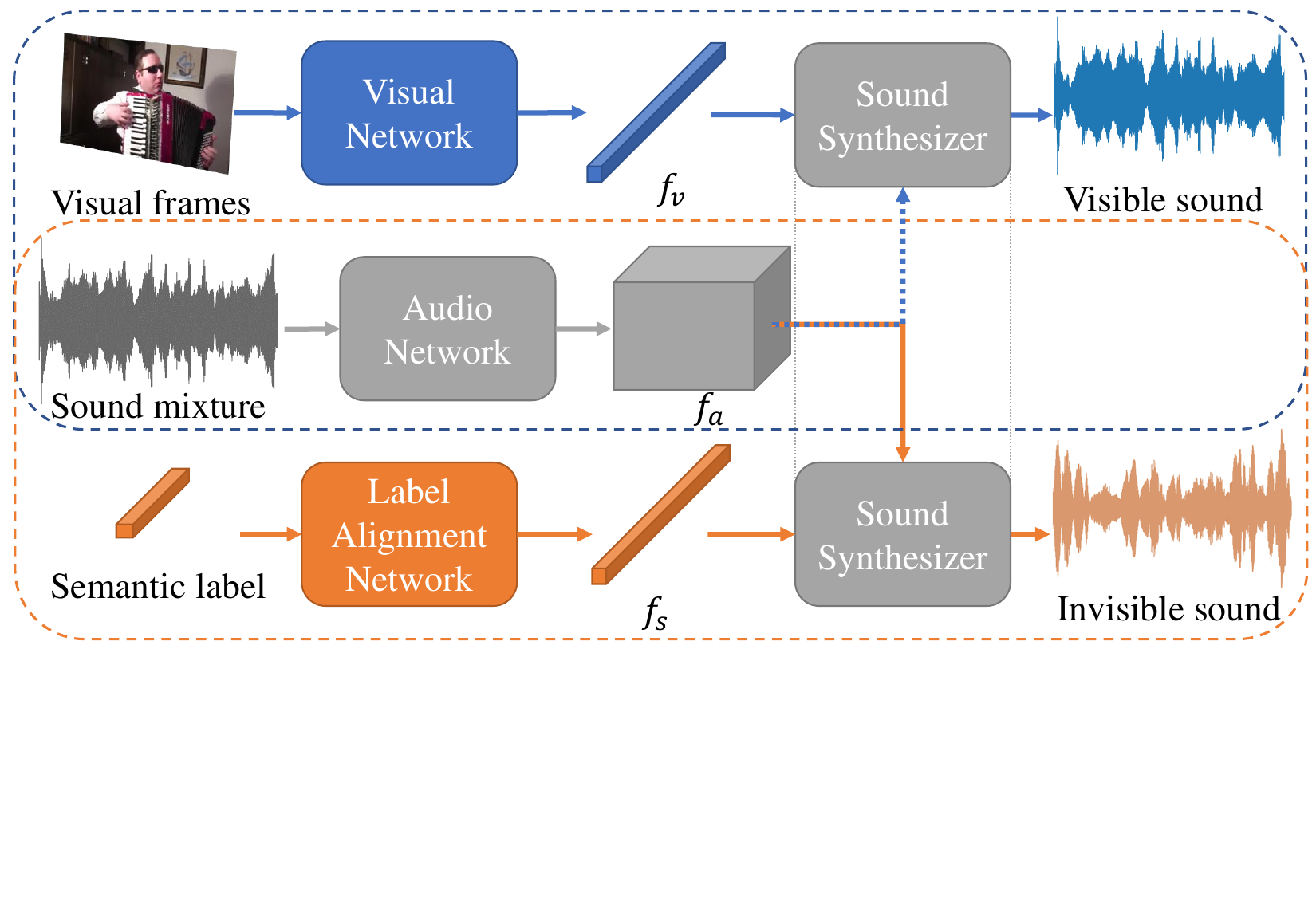}
    \vspace{-5em}
    \caption{The sound separators of AVSA-Sep. The audiovisual separator is in blue the semantic-guided separator is in orange, with shared components in grey. The visual network generates visual features $f_v$, the label alignment network produces aligned semantic label feature embeddings $f_s$, and the audio network yields audio features $f_a$ from the mixture spectrogram. Sound synthesizer predicts sound masks using audio and visual/semantic features and the same weights. Predicted masks combined with ISTFT recover predicted audios.}
    \label{fig:label_alignment}
\end{figure}

To address invisible sound sources, we suggest substituting semantic embeddings for visual features, shown in Fig.~\ref{fig:label_alignment}. This choice arises from their potential to guide sound separation, akin to visual features. To encode semantics, we introduce a label alignment network aligning semantic labels with visual features.

In the semantic alignment network, we encode a semantic label into a one-hot vector. A linear layer followed by a \verb|sigmoid| activation aligns this vector with visual features, yielding the semantic label embedding $f_s$ as the output.

The semantic alignment network introduces a semantic branch (\textit{semantic-guided separator}) alongside the visual branch (\textit{audiovisual separator}) in the sound separator. To ensure semantic embedding alignment with visual counterparts, we apply the same sound analysis network and sound synthesizer in both branches. 

\subsection{Audiovisual Scene Parser}

To achieve audiovisual scene-aware separation, we introduce a semantic parser. It predicts semantic labels for sound sources through audiovisual classification. Then, we apply the audiovisual sound separator and the semantic-guided separator for distinct source types.

\begin{figure}[tbp]
    \centering
    \includegraphics[width=.7\textwidth]{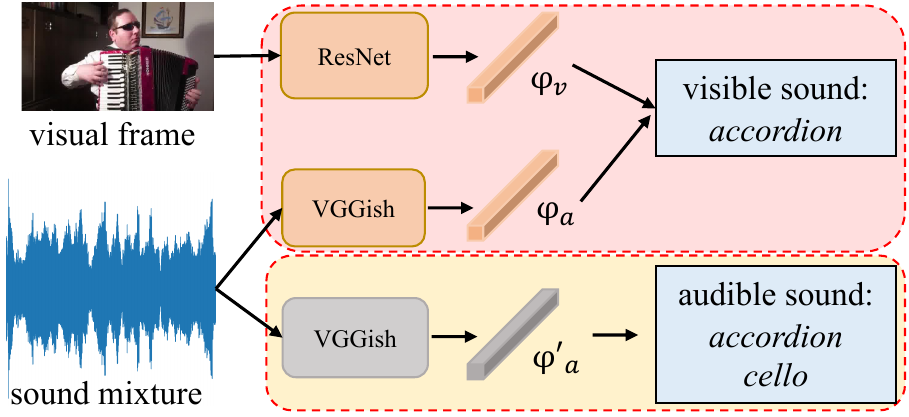}
    \caption{The figure shows the semantic parser. It employs a dilated ResNet-18 for visual features $\varphi_v$, and two VGGish networks for audio features $\varphi_a$ and $\varphi_a'$. Predicted semantic labels for visible sounds come from the fused $\varphi_v$ and $\varphi_a$, while audible sounds' labels originate from $\varphi_a'$.}
    \label{fig:parser}
\end{figure}

To cater to the semantic-guided separator, we predict both visible scene semantic labels $L_{\text{vis}}$ and semantic labels for inaudible but audible scenes $L_{\text{inv}}$. Visible sounds are those that are both audible and visible, while invisible sounds are audible but not visible. Our semantic parser, shown in Fig.~\ref{fig:parser}, comprises two main parts: the visible scene recognizer and the audible scene recognizer.

The visible parser includes a visual feature extractor, audio feature extractor, and fusion module merging audio and visual features. For video frames, a dilated ResNet-18 network generates $k_r$-channel visual features, pooled spatial-temporally. From audio mixture spectrograms, a VGGish network derives $k_r$-channel audio features through global pooling. The fusion module combines these features by summation. Fused features then transform semantic labels of visible scenes using a fully connected layer and \verb|sigmoid| activation.

The audible scene recognizer derives a $k_r$-channel feature from the audio mixture's spectrogram. Through a fully connected layer and \verb|sigmoid| activation, this feature transforms into semantic labels for audible scenes. It is important to note that the audio networks of the visible and audible scene recognizers have distinct weights due to differing tasks. The visible scene recognizer partners with the visual network to identify visible scenes, while the audible scene recognizer discerns both visible and invisible audio scenes.

\subsection{Joint Sound Separation Training}

During training, we independently train the semantic parser and sound separator, utilizing semantic labels for supervision and input, respectively. In inference, we merge the semantic parser and sound separator, enabling the whole model to predict both visible and invisible sounds from visual frames and sound mixture input.

The semantic parser is trained using a direct ''mix-and-predict'' approach. In each iteration, we mix sounds from randomly chosen videos, predict visible and audible sounds, and backpropagate loss. For AVSA-Sep and semantic parser training and evaluation, we adopt pipelines akin to those in Zhao~\etal~\cite{zhao_sound_2018}, with key modifications to optimize the joint training of the sound separator's two branches.

In each training and evaluation iteration, the sound separator processes a batch of $k$ videos. For each video, we mix audios from \(m-1\) randomly selected video clips to form the audio mixture \(S_{\text{mix}}\). The audiovisual separator utilizes frames from all \(m\) videos to predict individual sounds:
\(
\tilde{S}_1^{\text{vis}}, \tilde{S}_2^{\text{vis}}, \dots, \tilde{S}_m^{\text{vis}}.
\)
Similarly, the semantic-guided separator employs semantic labels of scenes from all \(m\) videos to predict individual sounds:
\(
\tilde{S}_1^{\text{scn}}, \tilde{S}_2^{\text{scn}}, \dots, \tilde{S}_m^{\text{scn}}.
\)

The loss for visual and semantic masks relative to ground truth masks is given by
\begin{align}
    \sum_{i=1}^{k} \text{loss}(S_i, \tilde{S}_i^{\text{vis}}) \text{ and } \sum_{i=1}^{k} \text{loss}(S_i, \tilde{S}_i^{\text{scn}}),
\end{align}
respectively. The overall sound separation loss is defined as
\begin{align}
    \mathcal{L}_{\text{ss}} = \sum_{i=1}^{k} \lambda \cdot \text{loss}(S_i, \tilde{S}_i^{\text{vis}}) + (2-\lambda) \cdot \text{loss}(S_i, \tilde{S}_i^{\text{scn}}),
\end{align}
where \(\lambda\) is a hyperparameter.
Additionally, we include the triplet loss among ground truth masks as an anchor, visual/semantic masks as positive, and semantic/visual masks of other sources as negative items, with coefficient $\eta$ as $\mathcal{L}_{\text{triplet}}= \sum_{i=1}^{k} \eta \cdot \text{triplet-loss}(S_i, \tilde{S}_i^{\text{vis}}, \tilde{S}_{-i}^{\text{scn}}))+\sum_{i=1}^{k} \eta \cdot \text{triplet-loss}(S_i, \tilde{S}_i^{\text{scn}}, \tilde{S}_{-i}^{\text{vis}}))$. The final loss becomes a combination of $\mathcal{L}_{\text{total}} = \mathcal{L}_{\text{ss}} + \mathcal{L}_{\text{triplet}}$.

These pipelines offer advantages such as joint weight updates during training and the combined evaluation of separation and scene recognition results against ground truth, yielding visual and semantic separation performance along with scene recognition performance.

%-------------------------------------------------------------------------
\section{Experiments}

\subsection{Experiment Setup}

\paragraph{Implementation Details.}

We implement our framework based on SoP~\cite{zhao_sound_2018}. Following their work, we choose to use binary masks and log-scale spectrograms. 
Our experiments use 6-second audio clips with three evenly spread frames per video. Feature channels are set at $k_r = 512$, and separation loss scaling is $\lambda = 1.5$. Sound separation quality is quantified using SDR and SIR metrics from \verb|mir_eval|~\cite{raffel2014mir_eval} to assess audiovisual and semantic-guided separation outcomes. 

\paragraph{Datasets.}
We trained and evaluated our model using the MUSIC dataset~\cite{zhao_sound_2018}, comprising 500 user-uploaded videos highlighting 11 musical instruments. Each instrument category includes around 50 videos. For training and validation, we excluded duet videos (15\%) lacking visible sounds. The dataset's clean and balanced nature is conducive to our task. We could generate artificial videos with invisible sounds for training and evaluation. Additionally, the dataset's accurate ground-truth scene labels are constructed from YouTube keyword queries, rendering it suitable for our purposes.

\subsection{Quantitative Results}

\begin{table}
\centering
% \resizebox{0.85\textwidth}{!}{
\begin{tabular}{llcccc}
\toprule
\multirow{2}{*}{\textbf{Dataset}} & \multirow{2}{*}{\textbf{Model}} & \multicolumn{2}{c}{\textbf{Visible}} & \multicolumn{2}{c}{\textbf{Invisible}} \\
\cmidrule(lr){3-4} \cmidrule(lr){5-6}
& & \textbf{SDR} & \textbf{SIR} & \textbf{SDR} & \textbf{SIR} \\ 
\midrule
\multirow{3}{*}{MUSIC} & Baseline \cite{zhao_sound_2018} & -0.65 & 6.06 & -1.92 & 0.24 \\
& MP-Net~\cite{xu2019recursive} & \textbf{-0.19} & 0.85 & -1.73 & -0.13 \\
& AVSA-Sep (Ours) & {-0.30} & \textbf{6.90} & \textbf{-1.41} & \textbf{5.43} \\
\bottomrule
\end{tabular}
% }
\caption{Comparison of our baseline (SoP~\cite{zhao_sound_2018}), MP-Net~\cite{xu2019recursive}, and AVSA-Sep (ours) in terms of visible/invisible SDR/SIR on the MUSIC dataset under the challenging 3-sound setting.}
\label{tab:3_mix_results}
\end{table}

In our experiments, we generate mixtures involving 3 sounds to introduce challenging scenarios with the potential for multiple invisible sounds. When assessing visible sounds, we compare our audiovisual separator's performance with that of existing audiovisual sound separators.

As invisible sound has no visual cue, conventional audiovisual sound separators cannot utilize frames associated with invisible sounds. These separators need to leverage all other frames to generate an output. Therefore, we follow these steps: We execute the established sound separation methods as usual and acquire the predicted sound components $\tilde{S}_1^{\text{vis}}$, $\tilde{S}_2^{\text{vis}}$, and $\tilde{S}_3^{\text{vis}}$. Then, we subtract the predicted sound for other sources,
\begin{align}
\tilde{S}_1^{\text{inv}} = S_{\text{mix}} - \tilde{S}_2^{\text{vis}} -\tilde{S}_3^{\text{vis}},
\end{align}
and similarly for $\tilde{S}_2^{\text{inv}}$ and $\tilde{S}_3^{\text{inv}}$, where $S_{\text{mix}}$ represents the input audio mixture.

This approach ensures that, during sound separation, audiovisual models disregard the associated frames for each sound, effectively treating both sounds as invisible. Consequently, the achieved metrics are comparable to those of the semantic-guided separator.

We present the results in Tab.~\ref{tab:3_mix_results}. The results show that AVSA-Sep outperforms the baseline (SoP~\cite{zhao_sound_2018}) and MP-Net~\cite{xu2019recursive} in terms of both visible and invisible sound separation quality, as indicated by higher SDR or SIR values. This highlights the effectiveness of the proposed AVSA-Sep model in handling both visible and invisible sound sources.

\subsection{Real-World Examples}

\begin{figure}
    \centering
    \includegraphics[width=0.77\textwidth]{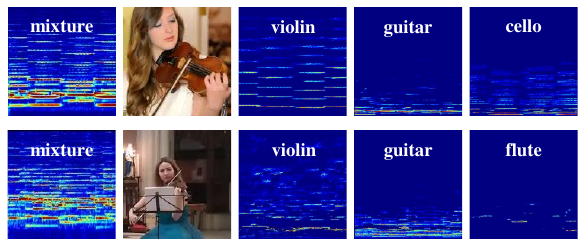}
    \caption{Two real-world trio videos. In both videos, only the violin is visible. Existing AVSS methods cannot separate the two invisible sounds. However, our approach can use semantic labels to separate them.}
    \label{fig:real_sep}
\end{figure}

We extend our evaluation beyond synthetic test cases to real-world video clips. To achieve this, we curate YouTube videos featuring 3 sounds and crop them to show only one of the musical instruments as visible. This approach allows us to gather real-world videos containing multiple invisible sounds. An illustrative instance of such a video, featuring one visible and two invisible sounds, is presented in Fig.~\ref{fig:real_sep}.

In this example, our model uses the audiovisual separator to estimate visible sounds and the semantic-guided separator to estimate invisible sounds. Note that since AVSS methods assume all sounds are visible, they cannot separate any of the two invisible sounds.

Even though there is no ground-truth spectrogram for these videos, we can still see from predicted spectrograms that our model is capable of separating visible sounds and multiple invisible sounds.

%-------------------------------------------------------------------------
\subsection{Ablation Studies}

\begin{table}
\centering
% \resizebox{0.47\textwidth}{!}{
\begin{tabular}{lcccc}
\toprule
\multirow{2}{*}{\textbf{Model}} & \multicolumn{2}{c}{\textbf{Visible}} & \multicolumn{2}{c}{\textbf{Invisible}} \\
\cmidrule(lr){2-3} \cmidrule(lr){4-5}
& \textbf{SDR} & \textbf{SIR} & \textbf{SDR} & \textbf{SIR} \\
\midrule
AV-Sep / Baseline \cite{zhao_sound_2018} & 7.74 & 14.80 & - & - \\
\midrule
SG-Sep & - & - & 6.23 & 12.80 \\
AVSA-Sep on Train Set & 10.73 & 17.67 & 9.72 & 16.22 \\
AVSA-Sep w/ Train Frames & 9.06 & 16.92 & - & - \\
AVSA-Sep & 7.91 & 14.81 & 9.01 & 16.45 \\
\bottomrule
\end{tabular}
% }
\caption{Ablation study results on the MUSIC dataset. SG-Sep refers to only the semantic-guided separator. [``On Train Set'': results evaluated on the training set; ``w/ Train Frames'': test frames are replaced with training frames of the same category.]}
\label{tab:ablation}
\end{table}

\paragraph{Joint training.}
Given our dual-branch design, we explore the interactions between the two branches by training them separately. The results are summarized in Tab.~\ref{tab:ablation}. 

Joint training enhances the performance of both the semantic-guided and audiovisual separators. This improvement is likely attributed to the use of complementary information from audiovisual modalities and semantic labels.

\paragraph{The visual branch \vs the semantic branch.}
The results in Tab.~\ref{tab:ablation} indicate that the semantic branch outperforms the visual branch when both are present. One potential explanation is the semantic branch's access to ground-truth labels, which the visual branch lacks. However, the semantic branch, without visual frames, cannot capture individual variations in videos. This is supported by our observation that while the semantic-guided separator achieves superior metrics on the test set, audiovisual separation performs better on the training set.

To investigate this further, we conduct an experiment where we replace the frames of test videos with frames from the training videos having the same scene label. The results, as summarized in Tab.~\ref{tab:ablation}, reveal a significant performance improvement when replacing the frames. This strongly supports the hypothesis that the underperformance of the visual branch is attributed to its inability to capture the semantic content in the frames of the test videos.

%------------------------------------------------------------------------
\section{Conclusion and Limitations}

\paragraph{Conclusion.}
We address the challenge of separating invisible sounds in AVSS and introduce a compatible framework with existing models. Our experiments confirm its capability to extend AVSS to multiple invisible sounds.

\paragraph{Limitations.}
While our approach handles multiple invisible sounds, exploring the treatment of multiple instances of the same semantic category could be valuable. Additionally, refining the semantic parser to enhance label prediction accuracy and potentially predict the counts of visible and invisible sounds represents another promising avenue.

\bibliographystyle{unsrtnat}
\bibliography{references}  %%% Uncomment this line and comment out the ``thebibliography'' section below to use the external .bib file (using bibtex) .

\begin{thebibliography}{55}
\providecommand{\natexlab}[1]{#1}
\providecommand{\url}[1]{\texttt{#1}}
\expandafter\ifx\csname urlstyle\endcsname\relax
  \providecommand{\doi}[1]{doi: #1}\else
  \providecommand{\doi}{doi: \begingroup \urlstyle{rm}\Url}\fi

\bibitem[Tian et~al.(2021)Tian, Hu, and Xu]{tian2021cyclic}
Yapeng Tian, Di~Hu, and Chenliang Xu.
\newblock Cyclic co-learning of sounding object visual grounding and sound separation.
\newblock In \emph{Proceedings of the IEEE/CVF Conference on Computer Vision and Pattern Recognition}, pages 2745--2754, 2021.

\bibitem[Liu et~al.(2022)Liu, Liu, Kong, Mei, Zhao, Huang, Plumbley, and Wang]{liu2022separate}
Xubo Liu, Haohe Liu, Qiuqiang Kong, Xinhao Mei, Jinzheng Zhao, Qiushi Huang, Mark~D Plumbley, and Wenwu Wang.
\newblock Separate what you describe: Language-queried audio source separation.
\newblock In \emph{INTERSPEECH}, 2022.

\bibitem[Tzinis et~al.(2022)Tzinis, Wisdom, Remez, and Hershey]{tzinis2022audioscopev2}
Efthymios Tzinis, Scott Wisdom, Tal Remez, and John~R Hershey.
\newblock Audioscopev2: Audio-visual attention architectures for calibrated open-domain on-screen sound separation.
\newblock In \emph{Proceedings of the European conference on computer vision (ECCV)}, 2022.

\bibitem[Zhu and Rahtu(2020)]{zhu2020visually}
Lingyu Zhu and Esa Rahtu.
\newblock Visually guided sound source separation using cascaded opponent filter network.
\newblock In \emph{Proceedings of the Asian Conference on Computer Vision (ACCV)}, November 2020.

\bibitem[Zhao et~al.(2018)Zhao, Gan, Rouditchenko, Vondrick, McDermott, and Torralba]{zhao_sound_2018}
Hang Zhao, Chuang Gan, Andrew Rouditchenko, Carl Vondrick, Josh McDermott, and Antonio Torralba.
\newblock The sound of pixels.
\newblock In \emph{Proceedings of the European conference on computer vision (ECCV)}, pages 570--586, 2018.

\bibitem[Afouras et~al.(2020)Afouras, Owens, Chung, and Zisserman]{Afouras20b}
Triantafyllos Afouras, Andrew Owens, Joon~Son Chung, and Andrew Zisserman.
\newblock Self-supervised learning of audio-visual objects from video.
\newblock In \emph{European Conference on Computer Vision}, 2020.

\bibitem[Gao and Grauman(2019)]{gao2019co}
Ruohan Gao and Kristen Grauman.
\newblock Co-separating sounds of visual objects.
\newblock In \emph{Proceedings of the IEEE International Conference on Computer Vision}, pages 3879--3888, 2019.

\bibitem[Majumder et~al.(2021)Majumder, Al-Halah, and Grauman]{majumder2021move2hear}
Sagnik Majumder, Ziad Al-Halah, and Kristen Grauman.
\newblock Move2hear: Active audio-visual source separation.
\newblock In \emph{Proceedings of the IEEE/CVF International Conference on Computer Vision}, pages 275--285, 2021.

\bibitem[Tzinis et~al.(2021)Tzinis, Wisdom, Jansen, Hershey, Remez, Ellis, and Hershey]{tzinis2020into}
Efthymios Tzinis, Scott Wisdom, Aren Jansen, Shawn Hershey, Tal Remez, Daniel P.~W. Ellis, and John~R. Hershey.
\newblock Into the wild with audioscope: Unsupervised audio-visual separation of on-screen sounds.
\newblock In \emph{International Conference on Learning Representations (ICLR) 2021}, 2021.
\newblock URL \url{https://arxiv.org/abs/2011.01143}.

\bibitem[Xu et~al.(2019)Xu, Dai, and Lin]{xu2019recursive}
Xudong Xu, Bo~Dai, and Dahua Lin.
\newblock Recursive visual sound separation using minus-plus net.
\newblock In \emph{Proceedings of the IEEE International Conference on Computer Vision}, pages 882--891, 2019.

\bibitem[Hyv{\"a}rinen and Oja(2000)]{hyvarinen2000independent}
Aapo Hyv{\"a}rinen and Erkki Oja.
\newblock Independent component analysis: algorithms and applications.
\newblock \emph{Neural networks}, 13\penalty0 (4-5):\penalty0 411--430, 2000.

\bibitem[Huang et~al.(2012)Huang, Chen, Smaragdis, and Hasegawa-Johnson]{huang2012singing}
Po-Sen Huang, Scott~Deeann Chen, Paris Smaragdis, and Mark Hasegawa-Johnson.
\newblock Singing-voice separation from monaural recordings using robust principal component analysis.
\newblock In \emph{2012 IEEE International Conference on Acoustics, Speech and Signal Processing (ICASSP)}, pages 57--60. IEEE, 2012.

\bibitem[Lee and Seung(2000)]{lee2000algorithms}
Daniel Lee and H~Sebastian Seung.
\newblock Algorithms for non-negative matrix factorization.
\newblock \emph{Advances in neural information processing systems}, 13, 2000.

\bibitem[Virtanen(2007)]{virtanen2007monaural}
Tuomas Virtanen.
\newblock Monaural sound source separation by nonnegative matrix factorization with temporal continuity and sparseness criteria.
\newblock \emph{IEEE transactions on audio, speech, and language processing}, 15\penalty0 (3):\penalty0 1066--1074, 2007.

\bibitem[Ozerov and F{\'e}votte(2009)]{ozerov2009multichannel}
Alexey Ozerov and C{\'e}dric F{\'e}votte.
\newblock Multichannel nonnegative matrix factorization in convolutive mixtures for audio source separation.
\newblock \emph{IEEE transactions on audio, speech, and language processing}, 18\penalty0 (3):\penalty0 550--563, 2009.

\bibitem[Fitzgerald et~al.(2016{\natexlab{a}})Fitzgerald, Liutkus, and Badeau]{fitzgerald2016projet}
Derry Fitzgerald, Antoine Liutkus, and Roland Badeau.
\newblock Projet—spatial audio separation using projections.
\newblock In \emph{2016 IEEE International Conference on Acoustics, Speech and Signal Processing (ICASSP)}, pages 36--40. IEEE, 2016{\natexlab{a}}.

\bibitem[Rickard(2007)]{rickard2007duet}
Scott Rickard.
\newblock The duet blind source separation algorithm.
\newblock In \emph{Blind speech separation}, pages 217--241. Springer, 2007.

\bibitem[Fitzgerald et~al.(2016{\natexlab{b}})Fitzgerald, Liutkus, and Badeau]{fitzgerald2016projection}
Derry Fitzgerald, Antoine Liutkus, and Roland Badeau.
\newblock Projection-based demixing of spatial audio.
\newblock \emph{IEEE/ACM Transactions on Audio, Speech, and Language Processing}, 24\penalty0 (9):\penalty0 1560--1572, 2016{\natexlab{b}}.

\bibitem[Hershey et~al.(2016)Hershey, Chen, Le~Roux, and Watanabe]{hershey2016deep}
John~R Hershey, Zhuo Chen, Jonathan Le~Roux, and Shinji Watanabe.
\newblock Deep clustering: Discriminative embeddings for segmentation and separation.
\newblock In \emph{2016 IEEE International Conference on Acoustics, Speech and Signal Processing (ICASSP)}, pages 31--35. IEEE, 2016.

\bibitem[Wang et~al.(2018)Wang, Le~Roux, and Hershey]{wang2018alternative}
Zhong-Qiu Wang, Jonathan Le~Roux, and John~R Hershey.
\newblock Alternative objective functions for deep clustering.
\newblock In \emph{2018 IEEE International Conference on Acoustics, Speech and Signal Processing (ICASSP)}, pages 686--690. IEEE, 2018.

\bibitem[Yu et~al.(2017)Yu, Kolb{\ae}k, Tan, and Jensen]{yu2017permutation}
Dong Yu, Morten Kolb{\ae}k, Zheng-Hua Tan, and Jesper Jensen.
\newblock Permutation invariant training of deep models for speaker-independent multi-talker speech separation.
\newblock In \emph{2017 IEEE International Conference on Acoustics, Speech and Signal Processing (ICASSP)}, pages 241--245. IEEE, 2017.

\bibitem[Kolb{\ae}k et~al.(2017)Kolb{\ae}k, Yu, Tan, and Jensen]{kolbaek2017multitalker}
Morten Kolb{\ae}k, Dong Yu, Zheng-Hua Tan, and Jesper Jensen.
\newblock Multitalker speech separation with utterance-level permutation invariant training of deep recurrent neural networks.
\newblock \emph{IEEE/ACM Transactions on Audio, Speech, and Language Processing}, 25\penalty0 (10):\penalty0 1901--1913, 2017.

\bibitem[Gao et~al.(2018)Gao, Feris, and Grauman]{gao2018learning}
Ruohan Gao, Rogerio Feris, and Kristen Grauman.
\newblock Learning to separate object sounds by watching unlabeled video.
\newblock In \emph{Proceedings of the European Conference on Computer Vision (ECCV)}, pages 35--53, 2018.

\bibitem[Zhao et~al.(2019)Zhao, Gan, Ma, and Torralba]{zhao_sound_2019}
Hang Zhao, Chuang Gan, Wei-Chiu Ma, and Antonio Torralba.
\newblock The sound of motions.
\newblock In \emph{Proceedings of the IEEE International Conference on Computer Vision}, pages 1735--1744, 2019.

\bibitem[Gan et~al.(2020)Gan, Huang, Zhao, Tenenbaum, and Torralba]{gan_music_2020}
Chuang Gan, Deng Huang, Hang Zhao, Joshua~B Tenenbaum, and Antonio Torralba.
\newblock Music gesture for visual sound separation.
\newblock In \emph{Proceedings of the IEEE/CVF Conference on Computer Vision and Pattern Recognition}, pages 10478--10487, 2020.

\bibitem[Rahman et~al.(2021)Rahman, Yang, and Sigal]{rahman2021tribert}
Tanzila Rahman, Mengyu Yang, and Leonid Sigal.
\newblock Tribert: Human-centric audio-visual representation learning.
\newblock In \emph{Thirty-Fifth Conference on Neural Information Processing Systems}, 2021.

\bibitem[Rahimi et~al.(2022)Rahimi, Afouras, and Zisserman]{rahimi2022reading}
Akam Rahimi, Triantafyllos Afouras, and Andrew Zisserman.
\newblock Reading to listen at the cocktail party: Multi-modal speech separation.
\newblock In \emph{Proceedings of the IEEE/CVF Conference on Computer Vision and Pattern Recognition}, pages 10493--10502, 2022.

\bibitem[Tan et~al.(2023)Tan, Ray, Burns, Plummer, Salamon, Nieto, Russell, and Saenko]{tan2023language}
Reuben Tan, Arijit Ray, Andrea Burns, Bryan~A Plummer, Justin Salamon, Oriol Nieto, Bryan Russell, and Kate Saenko.
\newblock Language-guided audio-visual source separation via trimodal consistency.
\newblock In \emph{Proceedings of the IEEE/CVF Conference on Computer Vision and Pattern Recognition}, pages 10575--10584, 2023.

\bibitem[Majumder and Grauman(2022)]{majumder2022active}
Sagnik Majumder and Kristen Grauman.
\newblock Active audio-visual separation of dynamic sound sources.
\newblock \emph{arXiv preprint arXiv:2202.00850}, 2022.

\bibitem[Tian et~al.(2020)Tian, Hu, and Xu]{tian2020co}
Yapeng Tian, Di~Hu, and Chenliang Xu.
\newblock Co-learn sounding object visual grounding and visually indicated sound separation in a cycle.
\newblock In \emph{IEEE/CVF Conference on Computer Vision and Pattern Recognition Workshops}, 2020.

\bibitem[Lee et~al.(2021{\natexlab{a}})Lee, Chung, Kim, Kang, and Sohn]{lee2021looking}
Jiyoung Lee, Soo-Whan Chung, Sunok Kim, Hong-Goo Kang, and Kwanghoon Sohn.
\newblock Looking into your speech: Learning cross-modal affinity for audio-visual speech separation.
\newblock In \emph{Proceedings of the IEEE/CVF Conference on Computer Vision and Pattern Recognition}, pages 1336--1345, 2021{\natexlab{a}}.

\bibitem[Gao and Grauman(2021)]{gao2021visualvoice}
Ruohan Gao and Kristen Grauman.
\newblock Visualvoice: Audio-visual speech separation with cross-modal consistency.
\newblock In \emph{2021 IEEE/CVF Conference on Computer Vision and Pattern Recognition (CVPR)}, pages 15490--15500. IEEE, 2021.

\bibitem[Montesinos et~al.(2022)Montesinos, Kadandale, and Haro]{montesinos2022vovit}
Juan~F. Montesinos, Venkatesh~S. Kadandale, and Gloria Haro.
\newblock Vovit: Low latency graph-based audio-visual voice sseparation transformer.
\newblock In \emph{European conference on computer vision}. Springer, 2022.

\bibitem[Chatterjee et~al.(2022)Chatterjee, Ahuja, and Cherian]{chatterjee2022learning}
Moitreya Chatterjee, Narendra Ahuja, and Anoop Cherian.
\newblock Learning audio-visual dynamics using scene graphs for audio source separation.
\newblock \emph{Advances in Neural Information Processing Systems}, 35:\penalty0 16975--16988, 2022.

\bibitem[Chatterjee et~al.(2021)Chatterjee, Le~Roux, Ahuja, and Cherian]{chatterjee2021visual}
Moitreya Chatterjee, Jonathan Le~Roux, Narendra Ahuja, and Anoop Cherian.
\newblock Visual scene graphs for audio source separation.
\newblock In \emph{Proceedings of the IEEE/CVF International Conference on Computer Vision}, pages 1204--1213, 2021.

\bibitem[Owens and Efros(2018)]{owens_audio-visual_2018}
Andrew Owens and Alexei~A Efros.
\newblock Audio-visual scene analysis with self-supervised multisensory features.
\newblock In \emph{Proceedings of the European Conference on Computer Vision (ECCV)}, pages 631--648, 2018.

\bibitem[Chen et~al.(2023)Chen, Zhang, Lian, Yang, Zeng, and Shi]{chen2023iquery}
Jiaben Chen, Renrui Zhang, Dongze Lian, Jiaqi Yang, Ziyao Zeng, and Jianbo Shi.
\newblock iquery: Instruments as queries for audio-visual sound separation.
\newblock In \emph{Proceedings of the IEEE/CVF Conference on Computer Vision and Pattern Recognition}, pages 14675--14686, 2023.

\bibitem[Gemmeke et~al.(2017)Gemmeke, Ellis, Freedman, Jansen, Lawrence, Moore, Plakal, and Ritter]{gemmeke2017audio}
Jort~F Gemmeke, Daniel~PW Ellis, Dylan Freedman, Aren Jansen, Wade Lawrence, R~Channing Moore, Manoj Plakal, and Marvin Ritter.
\newblock Audio set: An ontology and human-labeled dataset for audio events.
\newblock In \emph{2017 IEEE International Conference on Acoustics, Speech and Signal Processing (ICASSP)}, pages 776--780. IEEE, 2017.

\bibitem[Vedaldi et~al.(2020)Vedaldi, Zisserman, Chen, and Xie]{vedaldi2020vggsound}
A~Vedaldi, A~Zisserman, H~Chen, and W~Xie.
\newblock Vggsound: a large-scale audio-visual dataset.
\newblock In \emph{Proceedings of the International Conference on Acoustics, Speech, and Signal Processing}. IEEE, 2020.

\bibitem[Lee et~al.(2021{\natexlab{b}})Lee, Chung, Yu, Kim, Breuel, Chechik, and Song]{lee2021acav100m}
Sangho Lee, Jiwan Chung, Youngjae Yu, Gunhee Kim, Thomas Breuel, Gal Chechik, and Yale Song.
\newblock Acav100m: Automatic curation of large-scale datasets for audio-visual video representation learning.
\newblock In \emph{Proceedings of the IEEE/CVF International Conference on Computer Vision}, pages 10274--10284, 2021{\natexlab{b}}.

\bibitem[Lin et~al.(2023)Lin, Sung, Lei, Bansal, and Bertasius]{lin2023vision}
Yan-Bo Lin, Yi-Lin Sung, Jie Lei, Mohit Bansal, and Gedas Bertasius.
\newblock Vision transformers are parameter-efficient audio-visual learners.
\newblock In \emph{Proceedings of the IEEE/CVF Conference on Computer Vision and Pattern Recognition}, pages 2299--2309, 2023.

\bibitem[Gao et~al.(2023)Gao, Chen, and Xu]{gao2023collecting}
Junyu Gao, Mengyuan Chen, and Changsheng Xu.
\newblock Collecting cross-modal presence-absence evidence for weakly-supervised audio-visual event perception.
\newblock In \emph{Proceedings of the IEEE/CVF Conference on Computer Vision and Pattern Recognition}, pages 18827--18836, 2023.

\bibitem[Gong et~al.(2022)Gong, Rouditchenko, Liu, Harwath, Karlinsky, Kuehne, and Glass]{gong2022contrastive}
Yuan Gong, Andrew Rouditchenko, Alexander~H Liu, David Harwath, Leonid Karlinsky, Hilde Kuehne, and James~R Glass.
\newblock Contrastive audio-visual masked autoencoder.
\newblock In \emph{The Eleventh International Conference on Learning Representations}, 2022.

\bibitem[Zhou et~al.(2022)Zhou, Wang, Zhang, Sun, Zhang, Birchfield, Guo, Kong, Wang, and Zhong]{zhou2022audio}
Jinxing Zhou, Jianyuan Wang, Jiayi Zhang, Weixuan Sun, Jing Zhang, Stan Birchfield, Dan Guo, Lingpeng Kong, Meng Wang, and Yiran Zhong.
\newblock Audio--visual segmentation.
\newblock In \emph{European Conference on Computer Vision}, pages 386--403. Springer, 2022.

\bibitem[Mercea et~al.(2022{\natexlab{a}})Mercea, Hummel, Koepke, and Akata]{mercea2022temporal}
Otniel-Bogdan Mercea, Thomas Hummel, A~Sophia Koepke, and Zeynep Akata.
\newblock Temporal and cross-modal attention for audio-visual zero-shot learning.
\newblock In \emph{European Conference on Computer Vision}, pages 488--505. Springer, 2022{\natexlab{a}}.

\bibitem[Lee et~al.(2022{\natexlab{a}})Lee, Park, and Ro]{lee2022audio}
Sangmin Lee, Sungjune Park, and Yong~Man Ro.
\newblock Audio-visual mismatch-aware video retrieval via association and adjustment.
\newblock In \emph{European Conference on Computer Vision}, pages 497--514. Springer, 2022{\natexlab{a}}.

\bibitem[Cheng et~al.(2022)Cheng, Liu, Zhou, Qian, Wu, and Wang]{cheng2022joint}
Haoyue Cheng, Zhaoyang Liu, Hang Zhou, Chen Qian, Wayne Wu, and Limin Wang.
\newblock Joint-modal label denoising for weakly-supervised audio-visual video parsing.
\newblock In \emph{European Conference on Computer Vision}, pages 431--448. Springer, 2022.

\bibitem[Lee et~al.(2022{\natexlab{b}})Lee, Kim, and Ro]{lee2022weakly}
Sangmin Lee, Hyung-Il Kim, and Yong~Man Ro.
\newblock Weakly paired associative learning for sound and image representations via bimodal associative memory.
\newblock In \emph{Proceedings of the IEEE/CVF Conference on Computer Vision and Pattern Recognition}, pages 10534--10543, 2022{\natexlab{b}}.

\bibitem[Zellers et~al.(2022)Zellers, Lu, Lu, Yu, Zhao, Salehi, Kusupati, Hessel, Farhadi, and Choi]{zellers2022merlot}
Rowan Zellers, Jiasen Lu, Ximing Lu, Youngjae Yu, Yanpeng Zhao, Mohammadreza Salehi, Aditya Kusupati, Jack Hessel, Ali Farhadi, and Yejin Choi.
\newblock Merlot reserve: Neural script knowledge through vision and language and sound.
\newblock In \emph{Proceedings of the IEEE/CVF Conference on Computer Vision and Pattern Recognition}, pages 16375--16387, 2022.

\bibitem[Vasudevan et~al.(2022)Vasudevan, Dai, and Van~Gool]{vasudevan2022sound}
Arun~Balajee Vasudevan, Dengxin Dai, and Luc Van~Gool.
\newblock Sound and visual representation learning with multiple pretraining tasks.
\newblock In \emph{Proceedings of the IEEE/CVF Conference on Computer Vision and Pattern Recognition}, pages 14616--14626, 2022.

\bibitem[Afouras et~al.(2022)Afouras, Asano, Fagan, Vedaldi, and Metze]{afouras2022self}
Triantafyllos Afouras, Yuki~M Asano, Francois Fagan, Andrea Vedaldi, and Florian Metze.
\newblock Self-supervised object detection from audio-visual correspondence.
\newblock In \emph{Proceedings of the IEEE/CVF Conference on Computer Vision and Pattern Recognition}, pages 10575--10586, 2022.

\bibitem[Mercea et~al.(2022{\natexlab{b}})Mercea, Riesch, Koepke, and Akata]{mercea2022audio}
Otniel-Bogdan Mercea, Lukas Riesch, A~Koepke, and Zeynep Akata.
\newblock Audio-visual generalised zero-shot learning with cross-modal attention and language.
\newblock In \emph{Proceedings of the IEEE/CVF conference on computer vision and pattern recognition}, pages 10553--10563, 2022{\natexlab{b}}.

\bibitem[Mo and Tian(2022)]{mo2022multi}
Shentong Mo and Yapeng Tian.
\newblock Multi-modal grouping network for weakly-supervised audio-visual video parsing.
\newblock \emph{Advances in Neural Information Processing Systems}, 35:\penalty0 34722--34733, 2022.

\bibitem[Mittal et~al.(2022)Mittal, Morgado, Jain, and Gupta]{mittal2022learning}
Himangi Mittal, Pedro Morgado, Unnat Jain, and Abhinav Gupta.
\newblock Learning state-aware visual representations from audible interactions.
\newblock \emph{Advances in Neural Information Processing Systems}, 35:\penalty0 23765--23779, 2022.

\bibitem[Raffel et~al.(2014)Raffel, McFee, Humphrey, Salamon, Nieto, Liang, Ellis, and Raffel]{raffel2014mir_eval}
Colin Raffel, Brian McFee, Eric~J Humphrey, Justin Salamon, Oriol Nieto, Dawen Liang, Daniel~PW Ellis, and C~Colin Raffel.
\newblock mir\_eval: A transparent implementation of common mir metrics.
\newblock In \emph{In Proceedings of the 15th International Society for Music Information Retrieval Conference, ISMIR}. Citeseer, 2014.

\end{thebibliography}





% \end{thebibliography}

\end{document}